# An Exploration for higher efficiency in multi objective optimisation with reinforcement learning

Mehmet Emin AYDIN[1,2]

[1] University of the West of England, School of Computing and Creative Technologies, Bristol, UK
mehmet.aydin@uwe.ac.uk

[2] Istanbul Ticaret University, Dept. of Industrial Engineering, Istanbul, Türkiye
meaydin@ticaret.edu.tr

**ABSTRACT**

Efficiency in optimisation and search processes persists to be one of the challenges, which affects the performance and use of optimisation algorithms. Utilising a pool of operators instead of a single operator to handle move operations within a neighbourhood remains promising, but an optimum or near optimum sequence of operators necessitates further investigation. One of the promising ideas is to generalise experiences and seek how to utilise it. Although numerous works are done around this issue for single objective optimisation, multi-objective cases have not much been touched in this regard. A generalised approach based on multi-objective reinforcement learning approach seems to create remedy for this issue and offer good solutions. This paper overviews a generalisation approach proposed with certain stages completed and phases outstanding that is aimed to help demonstrate the efficiency of using multi-objective reinforcement learning.

## 1  INTRODUCTION

This paper introduces a generalisation approach through reinforcement learning in order to suggest a highly efficient swarm intelligence-based problem solver for combinatorial optimisation problems. It partially plays the role of a position paper but partially demonstrates completed work stages. Developing a general problem solver is one of the original targets of AI studies that attracts researcher since the golden time of artificial intelligence. However, the practice and studies in the field suggest that it is either extremely difficult to devise such a general problem solver or impossible [1].

Combinatorial optimisation problems are known as too difficult optimisation problems with classification of NP-Hard and/or NP-Complete problems. This escalates challenges into formulating efficient problem solver algorithms. Recently, heuristic-based approaches such as swarm intelligence or evolutionary algorithms have been studied extensively to ease these difficulties. One of the approaches emerged and proven success is the use of multiple operators within the population-based algorithms to keep the search diversified without losing intensification [2]. Reinforcement learning has taken researchers attention in building efficient rules to let adaptively select the most suitable operators subject to the search circumstances [3], [4]. Although the overall idea has been studied towards a mature level, no approach has appeared considering and examining the problem from generalisation of experiences point of view.

This paper investigates how an efficient approach can be devised to generalise the gained experiences in solving particular cases to utilise in different variants, similar use cases, and non-similar combinatorial problem types in such a way that gaining experience while solving a typical combinatorial problem (e.g. traveller salesman problem) and utilise them in solving very different problems such as job scheduling. For this purpose, the approach introduced here has two phases: (i) binarification of the problems, and (ii) transfer learning. By the means of binary representation all problems can be translated into a common space. Once binarification is completed and the suitability and usefulness of the binary operators in producing neighbouring solutions is sufficiently characterised, then a well-devised and customised machine learning algorithm such as reinforcement learning can help learn the characteristics of the operators in this respect to lead to generalisation of experiences.

This paper is organised as follows. Section 2 overviews the background of the study and the foundation knowledge in this respect covering the relevant work. Section 3 elaborates how operator selection can be conducted adaptively using reinforcement learning and how that drives to generalisation of experience. Section 4 present some proof-of-concept experimental studies, while Section 5 concludes the findings and future direction of this study.

## 2 BACKGROUND

Adaptive operator selection is a persisting problem in optimization domain with which the algorithms embedded with are studied for higher performance. Optimization problems are difficult and crucial problems, especially in engineering domains, and in the rest of the real-world that require robust, sustaining and reasonably quick solutions, while each of these objectives proportionately contributes towards the difficulty. Enormous studies have been conducted so far in order to ease the difficulties and challenges. Heuristic optimisation has emerged as non-traditional approaches spanning a vast horizon including many population-based approaches such as swarm intelligence algorithms. In this section, the foundation of adaptive operator selection problem will be introduced recognising the key contributions.

### 2.1 Optimisation Problem

Optimisation problem is structurally categorised as constrained and unconstrained problems. The majority of real-world problems are formulated as the constrained since the limited resources are required to be wisely used. A single objective optimisation problem is formulated as follows:

$$\begin{aligned}&\text{Minimize (or maximize) } f(x)\\&\text{subject to:}\\&\quad g_i(x) \leq 0 \quad \text{for } i = 1,..,m\\&\quad h_j(x) = 0 \quad \text{for } j = 1,..,p\\&\quad x \in X\end{aligned} \quad (1)$$

where $f(x)$ is the objective function, $g_i(x)$ and $h_j(x)$ are constraints, and $X$ defines the domain of variables. If $X$ is a discrete domain such as binary vectors, permutations, subsets, or integer vectors, then the optimisation problem will turn to be discrete or combinatorial optimisation. This is the definition of a single objective optimisation problems, where the quality of solutions is measured with objective function. On the other hand, the number of objectives increase to more than one, when the problem is attempted to be solved with respect to multiple quality measures. This means that a set of objectives, $f_i(x),$ where $i = 1,..,o$ is taken on board in measuring the quality of the solutions. The new form of optimisation problem with the set of functions turns to be as follows:

$$\begin{aligned}
&\text{Minimize (or maximize) } f_i(x), \; i = 1,..,o \\
&\text{subject to:} \\
&\quad g_j(x) \leq 0 \quad \text{for } j = 1,..,m \\
&\quad h_k(x) = 0 \quad \text{for } k = 1,..,p \\
&\quad x \in X
\end{aligned} \qquad (2)$$

where constraints and decision variables remain the same while objective functions are multiplexed.

This generic form is implemented based on domain details and relevant data to model the problems, which are solved with either of global [5] or local [6], [7] optimization algorithms. The local optimisation approaches include single solution and population driven algorithms. The study introduced in this paper considers population-driven heuristic optimisation algorithms including recently developed swarm intelligence algorithms such as artificial bee colony (ABC) optimisation [7], [8].

## 2.2 Multi-operator search and operator selection

Optimisation problems are solved with search algorithms in which problem states are perturbed with functions, $\varphi(x)$, in combinatorial cases, while a new solution is generated based on differentials, $\Delta x$, and derivatives, $\partial x$, of the solution in hand. The traditional search algorithms usually use single operators and devise the algorithmic structures accordingly. However, multiple operator choices have also been considered by heuristic search algorithms such as genetic algorithms with classical genetic operators of crossover and mutation. Crossover operator assists to move from one neighbourhood to another within the larger search space, while mutation is used to dive further within the same local region. The efficiency is handled via probabilities [9], [10]. Likewise, variable neighbourhood search organises shake and local search steps in a rather periodical structure, where the shake operator helps change the neighbourhood, while local search applies fine-tuning operations within a particular neighbourhood [10]. This idea clearly diversifies the search in order to find the most fruitful region of search space to reach the optimum or the best near optimum within shorter time. The diversification has been studied with memetic algorithms in which the population-based algorithms, especially evolutionary algorithms, are embedded with local search to customise the algorithms to solve problems in hand more efficiently.

The approach of regular or periodical use of 2 different operators has been extended with recruiting multiple operators or pools of operators managed with a tailored selection schemes to achieve diversification. The idea is to create a pool of operators, $\mathcal{O} = \{\varphi_i(x) | i = 1,..,|\mathcal{O}|\}$, and apply each to operate on the solutions orchestrated by a selection scheme, $\mathcal{S}: \mathcal{O} \times X \mapsto \mathbb{Z}$. The selection scheme is operating as in $\mathcal{S}(x) = \arg\max_{\varphi_i(x) \in \mathcal{O}} Q$, where $Q = \{q_i | i = 1,..,|\mathcal{O}|\}$ and $q_i = \sum_{t=0}^{T} q_{i,t}$ in which the operators are accumulating credits through. The operators are selected one after another for more efficient diversification in search.

The generic idea of using pool of operators and a selection scheme alongside the algorithm designed to conduct search is sketched in Figure 1. As seen, an 'algorithm' applies a 'selection scheme' to identify the best operator expected to produce the most fruitful solution to move to. "Selection scheme" recommends an operator from the "operator pool" and then the 'algorithm' applies that operator and produces a new solution. The solution is evaluated via the objective functions, and "Evaluate" unit produces feedback for corresponding selection and forwards to "credit assignment" module to calculate relevant credit level; once done the "selection scheme" is updated, accordingly.

Operator selection schemes are the rules in which the operators are evaluated and nominated to apply. The baseline rule is a uniformly distributed random selection rule, where all operators have equal likelihood to be opted. The commonly used rules are known as probability matching, adaptive pursuit [11], and multi-armed bandit-based methods [12], [13] which are evaluating the candidate operators based on their credit-level gained over the successes throughout the search conducted to the time. These rules do not map the problems states to operators, which does not consider the dynamics of the search process and circumstances. Some studies including [3], [4], [14] demonstrated that reinforcement learning can help dynamically map the states to the operators taking the search circumstances on board.

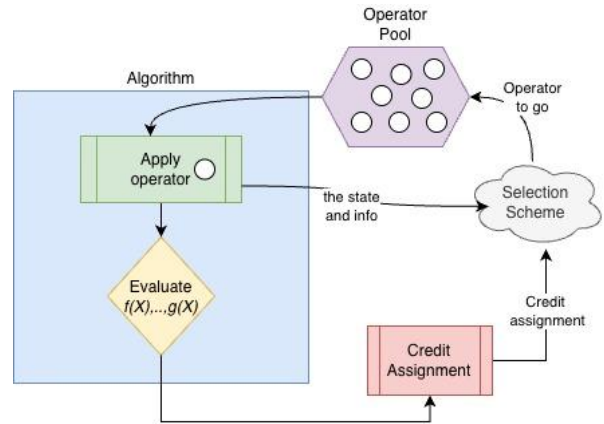

Figure 1: The process of search integrated with operator selection

### 2.3 Relevant work

Operator selection problem has been studied for decades with many respects to serve a number of trajectory-driven, also known as single solution-driven, and population-based algorithms [13], [15]. Recruitment of multiple operators in producing neighbours and alternative destination of search process to move to has emerged in various algorithms in different forms. As explained above, the purpose of using multiple operators is to diversify the search and avoid getting stuck in local optimum. It is expected that an operator can help moving the search from a persisting local neighbourhood. The most typical form is use of genetic operators in evolutionary algorithms, where *crossover*, *mutation* and *selection* operators are applied in a rather regular way, in which crossover operator is used to move from one neighbourhood of search space to another, while mutation helps dig-down to fine-tune the solutions in hand within the same neighbourhood [10], [16]. Likewise, variable neighbourhood search devices a regular structure to let the search change the neighbourhood with *shake* operator [10].

One of the approaches studied for diversification of the search without losing focus on intensification is the use of multiple operators in local search. Studies on adaptively activating mutation operators in evolutionary computing goes back decades; one of them can be found in [17]. Meanwhile, several studies were conducted to propose efficient operator selection schemes for multi-operator cases. Probability matching, adaptive pursuit and multi armed bandit-based approaches are among well-known ones [18], [19], [20], [21].

Reinforcement learning-based adaptive operator selection has also been taken on board and extensively studied [3], [4], [14], [22]. Although there are various studies appear in the literature paid attention on this problem, none of them have investigated the problem from a generalisation of experiences point of view. That justifies the idea and the relevant work plan to study ahead to obtain a clear view on how to generalise the experiences gained over search activities whilst recruiting reinforcement learning to improve the efficiency of adaptive operator selection and the algorithms recruit pools of operators and a bespoke selection scheme. The following section will introduce a generalisation approach based on reinforcement learning.

# 3 OPERATOR SELECTION WITH REINFORCEMENT LEARNING

## 3.1 Reinforcement learning

Reinforcement learning has been developed based on dynamic programming and Markov decision processes (MDP) in order to handle dynamically changing problems, where it is either hard to keep data to train machine learning models due to the temporal validity and/or dynamic nature of the problems. Due to this reason, the models are trained with running-time cases in which data can be generated and fed into the training process on time, while the data samples are not labelled, but the environment in which the acting agent is within reacts with signals. The returning signals are taken as positive or negative reward, which gives insight into the agent if the action taken was correct or not. This intricates the agent to learn what to do next given the circumstances.

The data structure of reinforcement learning is managed based on the 5-tuple of $(\mathcal{S}, \mathcal{A}, \mathcal{P}, \mathcal{R}, \gamma)$, $\mathcal{S}$ is the set of states, $\mathcal{A}$ is the set of actions that the agent is capable of acting, $\mathcal{P}$ is the transition probability, $\mathcal{R}$ is the reward that the agent will receive upon taking an action for a state, and $\gamma$ is the discount coefficient with which the accumulated rewards and normalised with. The main idea is that a reinforcement agent perceives a state $s \in \mathcal{S}$, and decides to take action $a \in \mathcal{A}$ and produces $s' \in \mathcal{S}$, with probability of $p = \mathcal{P}(s'|s, a)$. The environment returns a reward, $r = R(s, a)$, where $R \subseteq \mathcal{R}$. The agent makes a decision based on an evaluation criterion of $Q(s, a)$, which is provided with either through a deterministic or stochastic model. It repeats the same action forever ultimately aiming to maximise its accumulated reward, $R_t = \sum_{k=0}^{\infty} \gamma^k r_{t+k+1}$. Whenever the agent takes an action and receives a reward from its environment, the evaluation criterion $Q(s, a)$, for the pair of state, $s$ and action $a$ is updated following Bellman's equation applied to value iteration. It turns to be:

$$Q(s,a) \leftarrow Q(s,a) + \beta \left( r + \gamma \arg\max_{a' \in \mathcal{A}} Q(s',a') - Q(s,a) \right) \qquad (3)$$

where $\beta \in [0,1]$ and $\gamma \in [0,1]$ are known as learning rate and discount coefficient, while the expected criteria for the best winning action, $a'$ is identified with maximisation function.

## 3.2 Operator selection with reinforcement learning

Operator selection is studied to be conducted in a data-driven way in which machine learning and predictive models are proposed to take the role of selection rule in the process. Figure 2 sketches how the cycle of operator selection and crediting process work together, while a reinforcement learning process is inserted in following the 'Evaluate' operation as seen. The idea is to take up the evaluation further and processing the collected relevant data and information towards discovering knowledge patterns within the process. That would help detect when the problem state and the search circumstances turn to be suitable for operating with each operator function so that the process produces the best neighbouring solution to move to. The operator selection problem is attempted with a reinforcement learning model implemented with *Q learning* algorithm embedded with a modified *Hard C-means* algorithm to retain temporal differences-based experiences, [4]. First of all, the problem state is adopted as the environmental state perceived by the agent, $S \leftarrow \mathbf{x}$ where $S \subseteq \mathcal{S}$, and $\mathbf{x} \subseteq X$. Meanwhile the pool of operators is represented with $\mathcal{O}$, which is considered to be the set of actions, $\mathcal{A} \leftarrow \mathcal{O}$.

Figure 2 presents the logic that integrates reinforcement learning into the search cycle and shows how the credit assignment process for the selected operators works. As explained in the lefthand side section of Figure 2, the state of $\mathbf{x}$ is presented into operator selection while operator $o_i = \varphi_i(\mathbf{x})$ is provided as the selected one. The temporal difference between the state in hand, $\mathbf{x}$, and the generated one, $\mathbf{x}' = o_i(\mathbf{x})$, and the fitness values, $f(\mathbf{x})$ and $f(\mathbf{x}')$ are considered in evaluating of the success of the operator selection. As explained above, each operation contributes to the credit level of selected operator, while the credit level, $q_i$, is updated based on the success level achieved. Meanwhile, the reward function devised produces a reward value for the operation selection based on the success level that is considered in credit update of the selected operation. The recruited reinforcement learning algorithm for this purpose to handle credit assignment, $Q$ learning algorithm [4] is implemented in the way that the credit level of each operator is defined as, $q_i = Q(\mathbf{x}, o_i)$, where $\mathbf{x}$ and $o_i$ replace $s$ and $a$, in original $Q$ learning definition, respectively. Then, the new update function turns to be:

$$Q(\mathbf{x}, o_i) \leftarrow Q(\mathbf{x}, o_i) + \beta \left( r + \gamma \arg \max_{o_i \in \mathcal{O}} Q(\mathbf{x}', o_i) - Q(\mathbf{x}, o_i) \right) \quad (4)$$

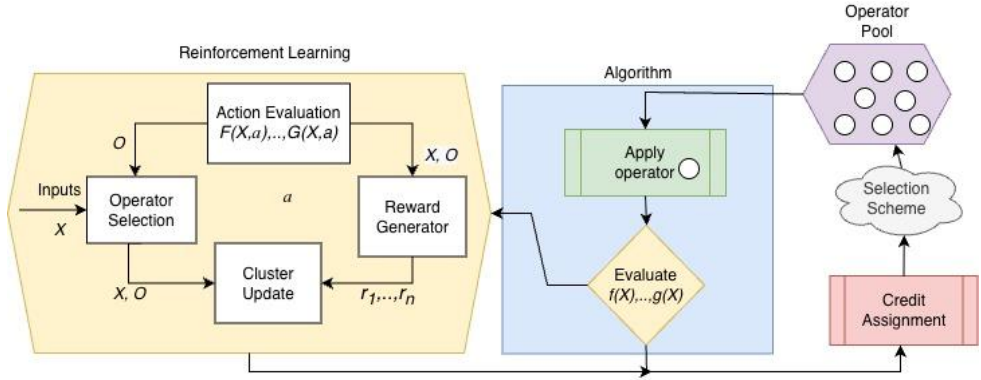

Figure 2: the search process integrated with RL-based operator selection

Meanwhile, the credit levels are preserved into a data model built based on *Hard-C-Means* algorithm [23], where each operator is represented as the centre of a cluster, $c_i \in C$ such that $C = \{c_i | i = 1, \ldots, |\mathcal{O}|\}$ and each cluster centre represents with the same data structure as $\mathbf{x}$, $|\mathbf{x}| = |c_i|$. The Euclidean distance in between $\mathbf{x}$ and $c_i$, $Q(\mathbf{x}, o_i) = \|\mathbf{x} - c_i\|$ is meant to be the credit level of the operator, $o_i$. After each successful operation, $c_i$ is updated with $c_i = \mathbf{x}/n_i$, and $n_i \leftarrow n_i + 1$, where $n_i$ is the counter of successful cases by the $i^{th}$ operator, $o_i$. This process is iterated until a predefined search process completed noting that $c_i$ will approximate to an optimum value after some while.

### 3.3 Operator selection for multi-objective optimisation

The previous subsection and Figure 2 describe how $Q$ learning is integrated into the multi-operator search algorithm in which the operators are adaptively selected and updated in credit for solving single objective combinatorial optimisation problems. Obviously, $arg \max_{o_i \in \mathcal{O}} Q(\mathbf{x}, o_i)$ is used for the decisions made over selection of operators in single optimisation cases, where each operator is represented with a single $Q$ value that reflects the preferability of the operator with respect to the only objective. However, the operators will be selected based on their credit scores with respect to each objective, when multi-objective optimisation problems have been considered for solutions in this way. Likewise, only one reward, $r$, is produced to reflect on the performance of the operator selected per move cycle. When the problem turns to be multi-objective, the reflection per selection has to be with respect to all objectives.

Let $\mathcal{F} = \{f_j(\mathbf{x}) | j = 1, \ldots, |\mathcal{F}|\}$ be the set of objective functions with which the decision variables, $\mathbf{x} \in X$ are aimed to be optimised with respect of each. The success of an operation with operator $o_i \in \mathcal{O}$ applied to a problem state $\mathbf{x} \in X$ to generate a neighbouring state, $\mathbf{x}' \in X$, would be assessed with respect to each objective, $\forall f_j(\mathbf{x}) \in \mathcal{F}$. The credit level of the chosen and applied operator, $o_i \in \mathcal{O}$, would be a set of $Q$ values, rather than a single value, which may be described with $Q_j = \{Q_j(\mathbf{x}, o_i) | j = 1, \ldots, |\mathcal{F}|\}$ for $\forall o_i \in \mathcal{O}$. Likewise, the reward estimated per operation, $r$, needs to be multiplexed in the same sense so that the operator $o_i \in \mathcal{O}$ would be rewarded as many as the number of objectives, $|\mathcal{F}|$. Therefore, each operation will receive a set of rewards, $r_i = \{R_j(\mathbf{x}, o_i) | j = $ $1, \ldots, |\mathcal{F}|\}$ for $\forall o_i \in \mathcal{O}$. After each operation, the credit level would be updated based on the following rule.

$$Q_j(\mathbf{x}, o_i) \leftarrow Q_j(\mathbf{x}, o_i) + \beta \left( R_j(\mathbf{x}, o_i) + \gamma \arg\max_{o_i \in \mathcal{O}} Q_j(\mathbf{x}', o_i) - Q_j(\mathbf{x}, o_i) \right), \forall f_j(\mathbf{x}) \in \mathcal{F} \quad (5)$$

Similar to the single-objective cases, the credit levels are preserved into a *Hard-C-Means –* based data model [23]. Each operator will be represented with a set of centres of a cluster, $c_{i,j} \in C$ such that $C = \{c_{i,j} | i = 1, \ldots, |\mathcal{O}|, \text{and } j = 1, \ldots, |\mathcal{F}|\}$ and each cluster centre represents with the same data structure as $\mathbf{x}$, $|\mathbf{x}| = |c_{i,j}|$. The Euclidean distance in between $\mathbf{x}$ and $c_{i,j}$, $Q_j(\mathbf{x}, o_i) = \|\mathbf{x} - c_{i,j}\|$ is meant to be the credit level of the operator, $o_i$. After each successful operation, $c_i$ is updated with $c_{i,j} = \mathbf{x}/n_{i,j}$, and $n_{i,j} \leftarrow n_{i,j} + 1$, where $n_{i,j}$ is the counter of successful cases by the $i^{th}$ operator, $o_i$ with respect to $j^{th}$ objective. This process is iterated until a predefined search process completed noting that $c_{i,j}$ will approximate to an optimum value after some while.

The key question is how to take on board the credits produced per objective in selecting an operator. The logic can be formulised as follows:

$$o_i = \arg\max_{o_i \in \mathcal{O}} \bigcup_{j=1}^{|\mathcal{O}|} Q_j(\mathbf{x}, o_i) \quad (6)$$

This union can be implemented simply through weighted sum of objectives, which unifies the objectives:

$$o_i = \sum_{j=1}^{|\mathcal{O}|} w_j Q_j(x, o_i) \quad (7)$$

where $w_j \in [0,1]$ and $\sum_{j=1}^{|\mathcal{O}|} w_j = 1$. However, this approach may end up with some reductions in the impact of each objective upon the decision to be made, which is an expected concern. Different implementation approaches (i.e., pareto front) are known in multi objective optimisation, which is applicable to this case, since this turns to be a multi-objective optimisation problem at this stage. Few recent studies including [24], [25] have been conducted in this regard, but there are many open questions outstanding.

## 4 EXPERIMENTAL RESULTS

The idea proposed in this article is an ongoing study in which single objective problem cases have been studied in various aspects but still more is outstanding. In addition, multi objective optimisation cases have not been experimented yet, therefore no results will be demonstrated in this regard. In the rest of this section, the experimental study conducted so far will be summarised to support the arguments and the essences of the proposal. The approach introduced in this article is part of an exploration journey towards a rather generic adaptive operator selection process integrated with relevant associated binary operators. It develops through multiple phases: (i) binary representation with binary operators, (ii) transfer learning across cases, and (iii) transfer learning across various combinatorial problems. Once a satisfactory level is achieved, multi objective optimisation problem cases will be taken on board.

### 4.1 Binary representation

The first step of proposed generalisation approach is to use binary representation of the problems states in hand and operate on them with binary operators. That has been conducted previously as in [4] to solve Set Union Knapsack Problem (SUKP) and demonstrated that the problems translated into binary space can work with binary operators borrowed from state-of-art studies as described in [11] and proposed by [18]. As plotted in Figure 3, the comparative results of three different operator selection rules embedded in ABC algorithm [8] to solve SUKP problem over 30 instances, where 3 different sizes are considered; each includes 10 variants. The selection rules to opt an operator per cycle of move are decided as 'Random' selection rule, 'PM' representing probability matching and 'RLABC' indicating a bespoke reinforcement learning algorithm. Obviously, the two algorithms other than 'RLABC' do not consider gaining and utilising the experience.

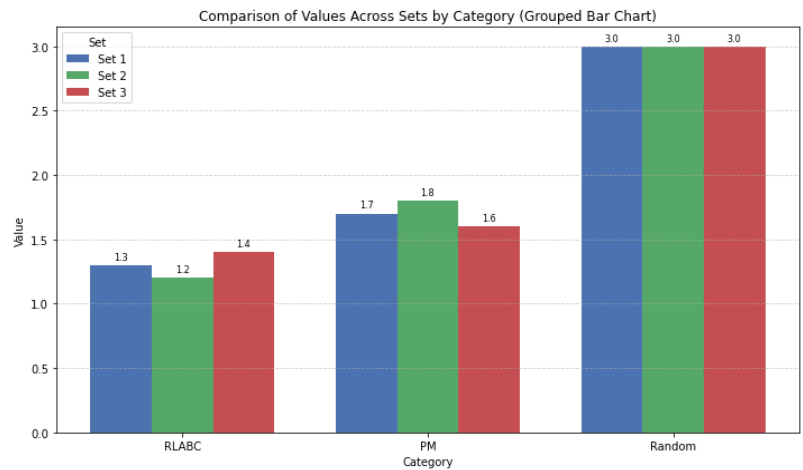

Figure 3: comparative results of ranks per operator selection rule embedded in ABC algorithm to solve SUKP problem instances.

In both [4], [11], more selection rules have been tested against one-another, where 'PM' is found to be best in many cases and competitive in many others. But, 'Random' select operators based on uniformly distributed random numbers. The results produced and displayed on Figure 3 are of Wilcoxon ranking sum test results suggesting the superiority of 'RLABC' over the other two, while "Random' remains the last all the time.

### 4.2 Transfer Learning

A transfer learning approach has been proposed by [26] in which the binary approach proposed in [4] has been utilised for solving very the same SUKP problem instances. 30 benchmarking SUKP instances in 3 different sizes – 10 instances in each size category. The results collected have been comparatively presented in various aspects, while the average ranking results (Wilcoxon ranking sum test results) of computational times have been plotted as in Figure 4, where RLABC is the stand-alone running RL embedded algorithm in ABC algorithm, which is used across these studies, while RLABC-T uploads previous experiences at the beginning, but does not apply continues learning. RLABC-TL upload previous experiences and keeps up with continues learning. The plots suggest that the best ranking is achieved by RLABC-TL, since it keeps using previous experience.

The type of transfer learning (transferring experiences) achieved in this experimentation is that RL is kept on supporting the search while repetitions of each benchmark are conducted. That helps solve the problems in shorter time in comparison to the cases switching off the RL over the reptations. However, a more sophisticate transfer learning across benchmarks in different sizes remain outstanding. More study has been conducted in this regard if a scalable approach to avoid size barrier in this approach, which is presented in both [27] and [28]. A set of latent features have been studied to represent the problem states alongside binary set. That has helped avoiding the scalability issue. But cross-problem transfer learning that is expected to allow utilisation of past experiences in solving new problems even they are in different type of combinatorial optimisation remain unsolved.

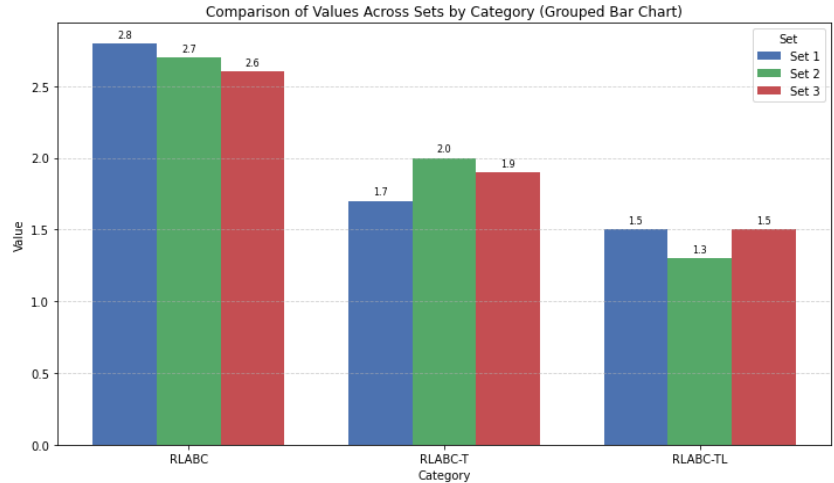

Figure 4: comparative results of ranking computed while comparing three versions of a RL-embedded ABC algorithm to solve SUKP problem instances.

### 4.3 Multi-objective optimisation problems

Multi-objective optimisation brings further levels of challenge in problem solving with transfer learning either across benchmarking instances or cross-problem level. The main outstanding issue to deal with is how to prioritise the objectives that help produce multiple rewards for very the same operation selection. Multi-objective reinforcement learning is required to be on board and implemented suitably and accordingly so that each operation would be related to every objective within the same context. The definition of the algorithms has been pencilled in the Section 3, while a thorough implementation is required to be conducted and validated.

## 5 CONCLUSION

In this study, a challenging problem of combinatorial optimisation handled with one of recent swarm intelligence algorithms, namely artificial bee colony (ABC), embedded with a reinforcement learning operator selection scheme is introduced. A generalisation approach to gain experience and utilise it across multiple layers is investigated. Cross-use case experience utilisation has been achieved while cross-problem type utilisation is outstanding. Also, multiple objective optimisation cases have not been much investigated with this point of view and many open questions outstand in this respect.